\documentclass{article}

\usepackage{PRIMEarxiv}

\usepackage[utf8]{inputenc} 
\usepackage[T1]{fontenc}    
\usepackage{hyperref}       
\usepackage{url}            
\usepackage{booktabs}       
\usepackage{amsfonts}       
\usepackage{nicefrac}       
\usepackage{microtype}      
\usepackage{lipsum}
\usepackage{fancyhdr}       
\usepackage{graphicx}       
\graphicspath{{media/}}     
\usepackage{amsmath}
\usepackage{multirow}

\title{ProtGO: A Transformer based fusion model for accurately predicting Gene Ontology (GO) terms from full scale Protein Sequences
\thanks{This work would be submitted to Scientific journals for possible publication. Copyright may be transferred without notice, after which this version may no longer be accessible.} 
}

\author{
  Azwad Tamir \\
  Department of ECE \\
  University of Central Florida \\
  Orlando\\
  \texttt{azwad.tamir@ucf.edu} \\
   \And
  Jiann-Shiun Yuan \\
  Department of ECE \\
  University of Central Florida \\
  Orlando\\
  \texttt{jiann-shiun.Yuan@ucf.edu} \\
}

\begin{document}
\maketitle

\begin{abstract}
Recent developments in next-generation sequencing technology have led to the creation of extensive, open-source protein databases consisting of hundreds of millions of sequences. To render these sequences applicable in biomedical applications, they must be meticulously annotated by wet lab testing or extracting them from existing literature. Over the last few years, researchers have developed numerous automatic annotation systems, particularly deep learning models based on machine learning and artificial intelligence, to address this issue. In this work, we propose a transformer-based fusion model capable of predicting Gene Ontology (GO) terms from full-scale protein sequences, achieving state-of-the-art accuracy compared to other contemporary machine learning annotation systems. The approach performs particularly well on clustered split datasets, which comprise training and testing samples originating from distinct distributions that are structurally diverse. This demonstrates that the model is able to understand both short and long term dependencies within the enzyme's structure and can precisely identify the motifs associated with the various GO terms. Furthermore, the technique is lightweight and less computationally expensive compared to the benchmark methods, while at the same time not unaffected by sequence length, rendering it appropriate for diverse applications with varying sequence lengths.
\end{abstract}



\section{Introduction}
The swift advancement of sequencing technologies has accelerated the growth of extensive protein sequence databases and made it both economically viable and practical to build large protein repositories both in the private domain and under open-source licenses. The substantial augmentation in the accessibility of protein sequences led to the establishment of protein databases such as Uniprot, with hundreds of millions of data points. Each day, about one hundred thousand proteins are added to the ever-expanding global public protein databases\cite{uniprot0}. To maximize the utility of these protein databases, they must be accurately annotated by the scientific community. Domain experts and curators exert considerable effort to manually annotate these protein sequences to address this issue; nonetheless, this procedure is exceedingly slow and resource intensive resulting in successful annotation of only approximately 0.03\% of proteins available in public databases\cite{uniprot0}.

The demand for expedited annotation of extensive protein datasets has led to the emergence of many automated and semi-automatic approaches. Various annotation systems have been developed over the years, employing distinct approaches to label proteins. One approach involved comparing the sequence of the target protein to other previously identified proteins to identify commonalities. The target received identical annotation to the known protein if substantial sequence segments between them exhibited alignment. BLAST\cite{blast} was among the most effective tools employing this methodology for annotating protein sequences.

A subsequent prevalent approach for protein annotation employed functional features to characterize target proteins. A protein is ascribed a certain function if it possesses the pattern linked to that function. Subsequent iterations of these models began employing profile Hidden Markov models (HMMs), enhancing their robustness and capacity to identify soft matches \cite{krogh, eddy}. These models calculated the likelihood of a new sequence possessing a specific function if the likelihood exceeded a set threshold. They are rapid and effective, resulting in numerous notable protein databases, such as Pfam and InterPro, utilizing them to annotate their extensive protein collections\cite{Interpro, pfam}. However, these models, while rapid and precise, possess certain limits.

Conventionally, they calculated the likelihood of each functional motif separately, resulting in reduced efficiency and accuracy when different functional groups shared a common characteristic. Furthermore, the signatures employed to identify a certain function in a protein sequence frequently necessitate human curator involvement and are not entirely automated, rendering the entire process quite tedious and laborious. These constraints the need for additional research in the area to develop more efficient annotation systems to manage the increasing volume of proteins that are continuously added to databases everyday.

Machine learning algorithms, particularly a variant known as deep learning characterized by models with multiple hidden layers, have garnered significant interest recently due to their success across various applications. Initial deep learning models focused on computer vision and machine translation applications, but they rapidly diversified into additional domains such as medical imaging and bioinformatics.

The functioning of a deep learning model typically comprises of two phases. The initial phase involves training, during which labeled and semi-labeled data is used to ascertain the model's weights, succeeded by the inference phase, where the acquired weights are employed to classify an unknown sample. The later layers of models, which engage in more generalized functions, may be utilized across analogous tasks. This approach is referred to as transfer learning, wherein, rather than training a model from scratch for a new task, the model's weights can be initialized from a comparable pretrained application. Subsequent to the transferring process, the model weights undergo training on the fresh data in a process referred to as fine-tuning. This method yields improved accuracy with reduced training data, requiring less training time and lower computational resources\cite{surveyTL}.

In light of the success of deep learning models across diverse domains and applications, numerous studies have been conducted to implement them in protein functional prediction and classification tasks\cite{deepGo, ECpred, ProLanGO, DeepLoc, Schwartz, DeePred, DeePre, DeepSF, HecNet1, HECNet2, allign_free, mlDEEPre}. Additionally, deep learning models have been applied to protein structure prediction\cite{mohammed, Senior2020, MSATrans, Yilun, Yang} and protein design\cite{Biswas2021, ProGen, Anishchenko2021, Yang2019, Mazurenko2020}.

A significant advancement in protein annotation commenced when researchers began inputting substantial volumes of unstructured raw data into extensive deep-learning models. This exempted the models from human biases, facilitated the identification of previously unrecognized patterns, and mitigated bottlenecks resulting from the time consuming curation process. Recent advancements in this domain encompass DeepEC\cite{DeepEC}, an ensemble Convolutional Neural Network (CNN) model that annotates protein sequences with Enzyme Commision (EC) numbers.

However, it can only annotate sequences shorter than one thousand units and necessitates a minimum of ten data points from a single class for optimal functionality. Other notable advancements in the area include CNN-based ensemble models for annotating unaligned protein domain sequences\cite{Bileschi2022}, the application of a transformer-based model to enhance both accuracy and computational efficiency\cite{Dohan}, the training of various models to comprehend protein sequence structures\cite{ProtTrans}, and the development of a transformer-based model for predicting protein functionality\cite{ProteinBert}. But these studies do not address whole protein sequences, which are frequently significant for bioinformatics applications. This issue has been addressed in Proteinfer\cite{Proteinfer}, which use both a singular and an ensemble CNN architecture to annotate whole protein sequences with GO terms and Enzyme Commission (EC) numbers.

We have previously developed a ProtBert\cite{ProtTrans} based model named ProtEC\cite{ProtEC} to accurately annotate EC numbers with high accuracy. This work further builds upon that to propose a classification based fusion model to predict Gene Ontology (GO) terms from full scale protein sequences that achieves state-of-the-art performance while at the same time addressing some of the major shortcomings of the previous models.

The primary novelties of our study are outlined below:
\begin{itemize}
  \item Achieving state-of-the-art accuracy on GO term annotation of full-scale protein sequences.
  \item Employing a singular model rather than an ensemble of models to enhance system efficiency and usability, while at the same time making it more light weight.
  \item The training time, GPU memory, and other computational resources needed to utilize the model are considerably lower compared to other similar methods.
  \item The proposed model features fully automated hyperparameter optimization, facilitating training without the need for a development or validation dataset.
  \item A superior comprehension of the protein sequence structure, which is demonstrated by the minimal accuracy disparity between the random and clustered dataset splits.
  \item Negligible dependency on the input sequence length, exhibiting robustness and the ability to be deployed in applications with significant input sequence length variability.
\end{itemize}

The paper is structured into four sections. The first section presents an overview of the study accompanied by appropriate literature reviews. The second section delineates the model architecture comprehensively and explains the training and evaluation techniques. The third section presents the experimental results of the proposed model and the performance comparison with other benchmark methods. Finally, the last section finishes the work and offers additional discussions on potential future research directions.

\section{Model}
This section first describes the input and output of the model, the dataset being used in this study, and the steps taken to preprocess the dataset. Next, the architecture of the proposed model is described in detail along with the training and evaluations steps and the experimental setup.

\subsection{GO Terms}
GO terms are keywords used to annotate protein sequences which contains various information like the gene that was used to transcribe the protein molecule, the type of organism which primarily uses the protein, the functionality of the protein, the biological processes and pathways connected to it, etc. The annotation of the protein sequence with GO terms is of utmost importance to the biological community as it helps identify the characteristics and properties of the protein along with its usage and functions within an organism. The GO terms could be divided into three categories based on the type of information they contain:
\begin{itemize}
    \item \textbf{Molecular Function:} Activities at the molecular level executed by gene products. Molecular function delineate processes that transpire at the molecular level, such as "catalysis" or "transportation." It denote activities rather than the entities (molecules or complexes) executing the actions, and do not delineate the location, timing, or context of the action. Molecular functions typically relate to activities executed by individual gene products (such as proteins or RNA), while certain activities are carried out by molecular complexes consisting of numerous gene products. Broad functional terms include catalytic activity and transporter activity, whereas narrower functional words encompass adenylate cyclase activity and Toll-like receptor binding.
    \item \textbf{Cellular Component:} This type of GO term describes the location occupied by a macromolecular apparatus relative to cellular compartments and structures. The gene ontology delineates the locations of gene products in two manners: the first method is by cellular anatomical entities, where a gene product performs a molecular function. Cellular anatomical entities encompass cellular components like the plasma membrane and cytoskeleton, along with membrane-bound compartments such as the mitochondrion. The other method is with the help of stable macromolecular complexes to which they belong, for instance, the clathrin complex.
    \item \textbf{Biological Process:} The extensive procedures, or 'biological programming,' executed through various molecular activities. Instances of general biological process terminology include DNA repair and signal transduction. Examples of narrower terminology include pyrimidine nucleobase biosynthesis process and glucose transmembrane transport.
\end{itemize}

\subsection{Dataset}
The protein sequences and their corresponding GO terms used in this study has been extracted from UniProt\cite{uniprot0}, which is a large public repository of protein sequences. The UniProt archive consists of two different databases; the first is called Swiss-Prot, which consists of about half a million protein sequences annotated manually by domain experts. The other database is known as TrEMBL and consists of around 250 million unreviewed sequences. The dataset used in this study has been prepared from the sequences taken from the SwissProt database as it contains reliable labels and could be used to train and test our model.

Two different dataset splits are used to train and evaluate our model. The first is the random split, where the training, development and testing sets are created by randomly splitting the whole dataset in a 8:1:1 ratio. The second dataset split is called the clustered split, where a tool called Uniref\cite{uniref} is used to divide the dataset into sections where the sequences in each splits are different in structure and has the least possible common segments between them. This makes the clustered split dataset much more challenging compared to the former. The number of datapoints in the random and clustered split datasets are given in Table~\ref{Table_random} and Table~\ref{Table_clustered1}, respectively.

\begin{table}
\renewcommand{\arraystretch}{1.5}
\setlength{\tabcolsep}{6.0pt}
\caption{Number of sequences of each aspect in the random dataset splits}
\label{Table_random}
\centering
\begin{tabular}{|c||c|c|c|c|}
\hline
\textbf{Split} & \textbf{Proteins} & \begin{tabular}[c]{@{}c@{}}\textbf{Biological}\\ \textbf{Processes}\end{tabular} & \begin{tabular}[c]{@{}c@{}}\textbf{Molecular} \\ \textbf{Functions}\end{tabular} & \begin{tabular}[c]{@{}c@{}}\textbf{Cellular} \\ \textbf{components}\end{tabular}\\
\hline\hline
\textbf{Train} & 438,522 & 346,677 & 369,909 & 321,980\\
\hline
\textbf{Development} & 55,453 & 43,734 & 46,785 & 40,765\\
\hline
\textbf{Test} & 54,289 & 42,830 & 45,662& 39,929\\
\hline
\end{tabular}
\end{table}


\begin{table}
\renewcommand{\arraystretch}{1.5}
\setlength{\tabcolsep}{6.0pt}
\caption{Number of sequences of each aspect in the clustered dataset splits}
\label{Table_clustered1}
\centering
\begin{tabular}{|c||c|c|c|c|}
\hline
\textbf{Split} & \textbf{Proteins} & \begin{tabular}[c]{@{}c@{}}\textbf{Biological}\\ \textbf{Processes}\end{tabular} & \begin{tabular}[c]{@{}c@{}}\textbf{Molecular} \\ \textbf{Functions}\end{tabular} & \begin{tabular}[c]{@{}c@{}}\textbf{Cellular} \\ \textbf{components}\end{tabular}\\
\hline\hline
\textbf{Train} & 182,965 & 144,292 & 154,150 & 134,200\\
\hline
\textbf{Development} & 180,309 & 143,130 & 152,156 & 131,778\\
\hline
\textbf{Test} & 183,475 & 144,605 & 155,593 & 135,345\\
\hline
\end{tabular}
\end{table}

Next, during the preprocessing steps, the sequences in the dataset that does not have any GO term annotations associated with them are filtered out. After that, the list of GO terms present in the dataset are collected and divided based on their aspects, which are biological processes, cellular components and molecular functions. Each dataset is next split into three parts where each contains only GO terms confined to one aspect. Only the top 100 most populous GO terms from each aspect are used for training and evaluation as many of them only appear in the dataset very rarely. The number of protein sequences in each aspect for the training, development and testing sets are also given in Table~\ref{Table_random} and Table~\ref{Table_clustered1} for the random and clustered datasets respectively.

After separating out the datasets based on the three aspects, the GO terms are one-hot encoded where, a value of 1 is assigned to the terms that are present for a particular protein sequence and 0 if they are absent. Next, the protein sequences are tokenized and inputted into the algorithm while the one-hot encoded GO terms serves as the truth labels or targets.

\begin{figure}[h]
  \centering
  \includegraphics[width=\linewidth]{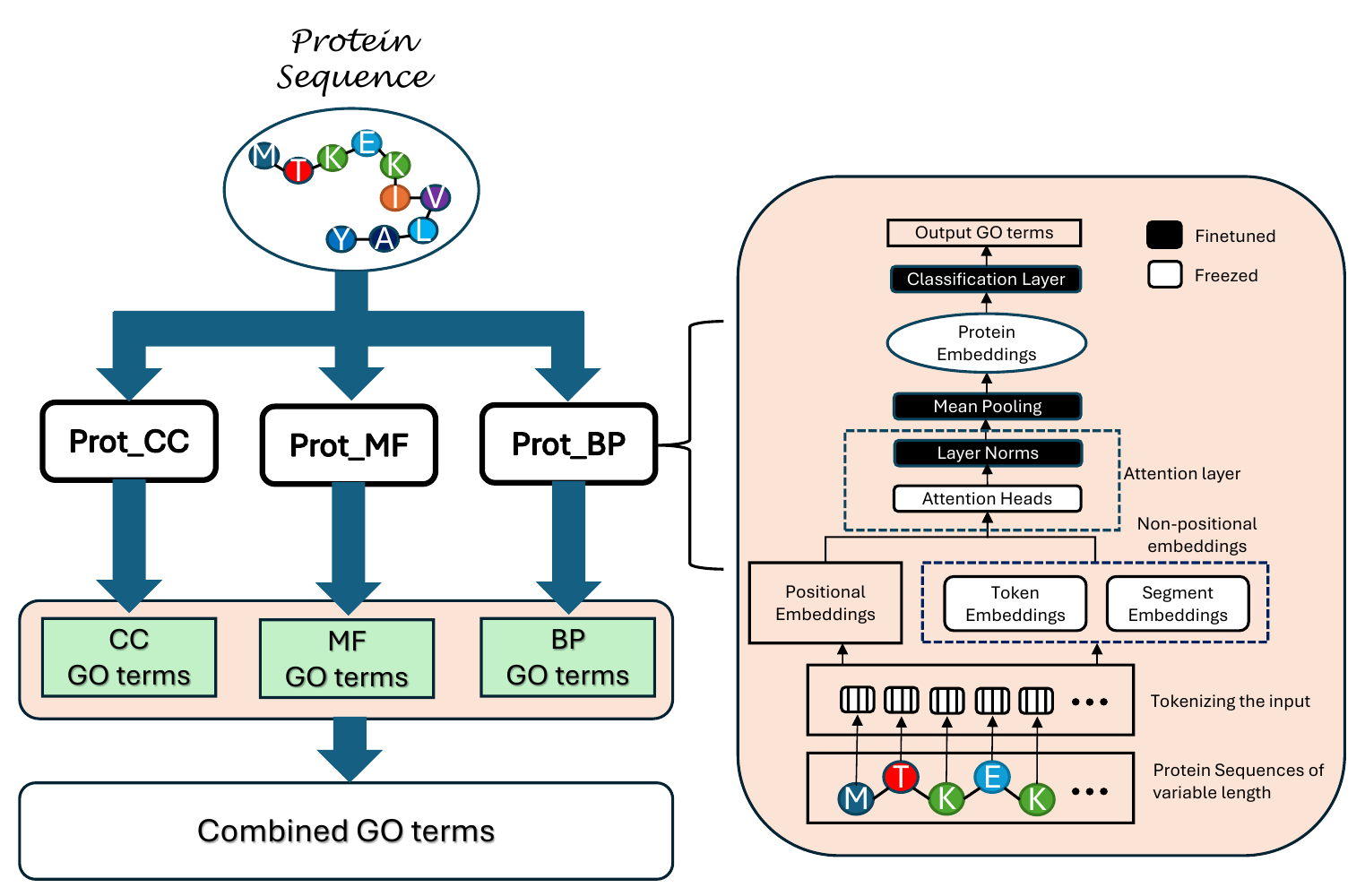}
  \caption{Block diagram illustrating the detailed architecture of the proposed model of ProtGO.}
  \label{Fig_model}
\end{figure}

\subsection{Model Architecture}
A block diagram illustrating the detailed architecture of the proposed model is given in Fig.\ref{Fig_model}. Three transformer based modules are used in the model, where each specializes on one aspect of the GO terms. The protein sequence is first tokenized, where each amino acid in the sequence is represented by an unique alphabet. The tokenizer takes the input sequence and converts it into a string of vectors which could be now inserted into the transformer architecture.

The protein sequence is copied and parallelly inserted into all three of the transformer blocks; Prot\_CC, Prot\_MF, and Prot\_BP. Each of these modules predict the GO terms related to one of the aspects. Next, the predictions from all three blocks are combined together to create the total output of the model.

The internal architecture of each of the prot\_X block is also given in Fig.\ref{Fig_model}. After the input protein sequences are tokenized, they are fed into three encoders namely, the positional encoder, the token encoder and the segment encoder. Each of the encoder is responsible for generating one aspect of the input tokens. The token encoder and the segment encoder are together referred to as non-positional encoders.

Internally, each of the encoders contain learnable weights which could be factored with the inputs to generate the output embeddings. The positional enbeddings contains information about the position of each token in the sequence, while the token embeddings describes the type of token in each position. Lastly, the segment embeddings mostly remains dormant in this case as the protein sequences are not divided into multiple segments. After all the three types of embeddings are generated, they are summed together and fed into the transformer encoder.

The encoder module of the transformer is made up of multiple encoder layers stacked on top of each other. The proposed model consists of 30 encoder layers, where each consists of a multihead attention module followed by layer normalization and mean pooling. Each encoder layer also consists of a skip connection or residual network that facilitates the propagation of early layer information to the later layers of the model. A model dimensionality of 1024 is used along with 12 heads for the attention layers. After the transformer encoder extracts the embeddings from the input sequence, they are fed into a classification layer which predicts the GO terms that the input sequence are associated with. Lastly, the labels from all three submodules are collected and pooled together to form the final model output.

\subsection{Training and Evaluation}
The first step in the training process involves pretraining the Prot\_X blocks on the BFD100 dataset which consists of 2,122 million protein sequences of variable lengths\cite{ProtTrans}. The pretrained variant of ProtBert was obtained from the HuggingFace website\cite{wolf2019huggingface}. During pretraining, segments of the sequence, determined by the masking probability, were concealed, and the model was instructed to predict the masked tokens. This enables the model to comprehend and assimilate intricate patterns within the data, as well as the interrelations among various amino acids in the protein sequence. The ProtBert module comprised of 30 hidden layers, each containing 1024 nodes, with an intermediate hidden layer size of 4096. The model includes 16 attention heads and contained a total of 420 million parameters. It was pretrained in \cite{ProtTrans} utilizing 128 nodes and 1024 TPUs for 800,000 steps on sequences limited to a maximum length of 512, followed by an additional 200,000 steps on sequences with a maximum length of 2,000. A masking probability of 15\% was implemented in conjunction with a Lamb optimizer. A learning rate of 0.002 and a weight decay rate of 0.01 were chosen.

To adapt the ProtBert module for our purposes, we acquired the pretrained weights and fine-tuned them on enzyme sequences. Nonetheless, fine-tuning substantial transformer models such as ProtBert demands significant data and computational resources. To address this, we have executed selective fine-tuning, wherein certain layers of the model are adjusted while others retain their pretrained values. This strategy enhances accuracy in large models with limited data availability\cite{UniversalEngines, AMP_Deep}. The fundamental architecture of the ProtBert model, along with the selectively fine-tuned layers, is illustrated in Fig. \ref{Fig_model}. An Negative Log Likelihood (NLL) loss function is used to pretrain the modules in an unsupervised manner. The layers which were frozen to get the most efficiency for the ProtBert modules were referenced from \cite{ProtEC}. 

Next, the pretrained Prot\_Bert modules are selectively finetuned on the processed random and clustered datasets. The training set from each type of split is used to train the model, while the testing set has been reserved for the evaluation and comparison with benchmarks methods. The cross entropy loss function has been used during the finetuning process and is given in equation 1. Here, N is the total number of samples in the dataset while M is the total number of classes, y represents the true label and \( \hat{y} \) stands for the predicted labels. An Adam optimizer has been used to train the model on a total of 10 epochs with an initial learning rate of 5e-4. The HuggingFace\cite{wolf2019huggingface} trainer has been used to train the model which automatically adjusts and optimizes most of the hyperparameters along with a gradient accumulation step of 32. The model was trained on a single Nvidia RTX2080Ti GPU which required around 3.5GB of memory and took approximately 80 hours to be completely trained.

After the training step, the model was evaluated on the held out testing set for both the random and clustered split datasets. The results obtained in the experiment was compared with the benchmark models and are presented in detail in the next section.

\[ Loss = -\frac{1}{N} \sum_{i}^N \sum_{j}^M y_{ij}log(\hat{y}_{ij}) \hspace{1.5cm} (1)\]

\section{Results}

This section presents the evaluation statistics and performance comparison of the proposed ProtGO model with previous state-of-the-art benchmark models. Other analysis results including the Receiver Operating Characteristic (ROC) curve and sequence length analysis of the proposed model have also been outlined and discussed in this section.

\begin{table}[]
\renewcommand{\arraystretch}{1.6}
\setlength{\tabcolsep}{7.5pt}
\caption{Evaluation results of the proposed model on the Random Split dataset on various performance metrics compared to the benchmark methods.}
\label{Table_Random}
\centering
\begin{tabular}{|c||c|c|c|c|c|}
\hline
\textbf{Aspect}                              & \textbf{Model}          & \textbf{Accuracy}         & \textbf{F1 score}        & \textbf{Precision}       & \textbf{Recall}          \\ \hline \hline
\multirow{3}{*}{\textbf{Biological Processes}} & \textbf{ProtGO} & \textbf{86.06\%} & \textbf{0.9251} & \textbf{0.9725} & 0.8821 \\ \cline{2-6} 
                                    & \textbf{Proteinfer}     & 80.21\%          & 0.8902          & 0.8447          & 0.9409          \\ \cline{2-6} 
                                    & \textbf{Proteinfer\_EN} & 83.29\%          & 0.9088          & 0.8652          & \textbf{0.9570} \\ \hline \hline
\multirow{3}{*}{\textbf{Molecular Function}} & \textbf{ProtGO}         & \textbf{94.60\%} & \textbf{0.9722} & \textbf{0.9882} & 0.9568          \\ \cline{2-6} 
                                    & \textbf{Proteinfer}     & 88.94\%          & 0.9415          & 0.9166          & 0.9677          \\ \cline{2-6} 
                                    & \textbf{Proteinfer\_EN} & 91.67\%          & 0.9565          & 0.9344          & \textbf{0.9797} \\ \hline \hline
\multirow{3}{*}{\textbf{Cellular Component}} & \textbf{ProtGO}         & \textbf{78.30\%} & \textbf{0.8783} & \textbf{0.9469} & 0.8189          \\ \cline{2-6} 
                                    & \textbf{Proteinfer}     & 69.36\%          & 0.8191          & 0.7386          & 0.9191          \\ \cline{2-6} 
                                    & \textbf{Proteinfer\_EN} & 75.40\%          & 0.8597          & 0.7928          & \textbf{0.9390} \\ \hline
\end{tabular}%
\end{table}

Several evaluation metrics have been used to comprehensively compare the results of the models on both the random and clustered dataset splits. Table \ref{Table_Clustered} shows the performance of the proposed model on all three GO aspects, namely, Biological Processes, Molecular Function, and Cellular Component on the random split dataset using the overall Accuracy, F1 score, Precision, and Recall evaluation metrics. The results show that the proposed model, ProtGO has been able to perform significantly better compared to the main benchmark methods Proteinfer and ProteinferEN\cite{Proteinfer}. Here, ProteinferEN refers to the ensembled version of the Proteinfer model, where several base models have been trained and later put together to generate the final output. 

ProtGO outperformed ProteinferEN by around 3\% in terms of Accuracy in all three GO terms aspects while showing better F1 scores as well. The Precision scores of ProtGO are also significantly better than the other methods. However, in terms of the Recall scores, ProteinferEN shows better results out of the three compared models. This delineates that the ProteinferEN model is better at avoiding false negative samples compared to ProtGO and vice versa. But the trade off between the false positive and false negative samples could be controlled using a minimum decision threshold hyperparameter, which has been set to the default value for all the algorithms. Hence, a better indicator of performance is the F1 score, which combines the values of Precision and Recall to produce a single metric. In terms of the F1 score, the ProtGO module has been able to consistently outperformed the benchmark methods in both the datasets.

\begin{table}[]
\renewcommand{\arraystretch}{1.6}
\setlength{\tabcolsep}{7.5pt}
\caption{Evaluation results of the proposed model on the Clustered Split dataset on various performance metrics compared to the benchmark methods.}
\label{Table_Clustered}
\centering
\begin{tabular}{|c||c|c|c|c|c|}
\hline
\textbf{Aspect}                                & \textbf{Model}  & \textbf{Accuracy} & \textbf{F1 score} & \textbf{Precision} & \textbf{Recall} \\ \hline \hline
\multirow{3}{*}{\textbf{Biological Processes}} & \textbf{ProtGO} & \textbf{82.16\%}  & \textbf{0.9021}   & \textbf{0.9532}    & 0.8561          \\ \cline{2-6} 
 & \textbf{Proteinfer}     & 72.48\% & 0.8424 & 0.8940 & 0.7965          \\ \cline{2-6} 
 & \textbf{Proteinfer\_EN} & 75.56\% & 0.8650 & 0.8618 & \textbf{0.8683} \\ \hline \hline
\multirow{3}{*}{\textbf{Molecular Function}}   & \textbf{ProtGO} & \textbf{91.51\%}  & \textbf{0.9556}   & \textbf{0.9785}    & \textbf{0.9338} \\ \cline{2-6} 
 & \textbf{Proteinfer}     & 81.51\% & 0.8992 & 0.9241 & 0.8756          \\ \cline{2-6} 
 & \textbf{Proteinfer\_EN} & 83.88\% & 0.9159 & 0.9012 & 0.9312          \\ \hline \hline
\multirow{3}{*}{\textbf{Cellular Component}}   & \textbf{ProtGO} & \textbf{73.28\%}  & \textbf{0.8458}   & \textbf{0.9215}    & 0.7817          \\ \cline{2-6} 
 & \textbf{Proteinfer}     & 63.06\% & 0.7782 & 0.8014 & 0.7563          \\ \cline{2-6} 
 & \textbf{Proteinfer\_EN} & 66.22\% & 0.8047 & 0.7832 & \textbf{0.8277} \\ \hline
\end{tabular}%
\end{table}

Table \ref{Table_Clustered} shows the performance comparison of the proposed model along with the benchmark methods on the clustered split dataset. The results are similar to the random split, with ProtGO performing better in terms of Accuracy and F1 score in all three GO aspects. However, the gap between the performance between ProtGO and the benchmark methods is more in case of the clustered split compared to the random split dataset. This shows that ProtGO is a more robust model and have a strong understanding of the structure and deep motif patterns of the protein structure. Hence, it performs similarly for the random and clustered test dataset while the benchmark models show significant difference in Accuracy for the two datasets. There is a gap of around 7\% in Accuracy between the proposed model and the benchmark methods for the clustered split dataset which is significantly greater than that seen for the random dataset split. A similar trend could also be seen in the F1 score, Precision and Recall scores for the two splits.

\begin{figure}[h]
  \centering
  \includegraphics[width=\linewidth]{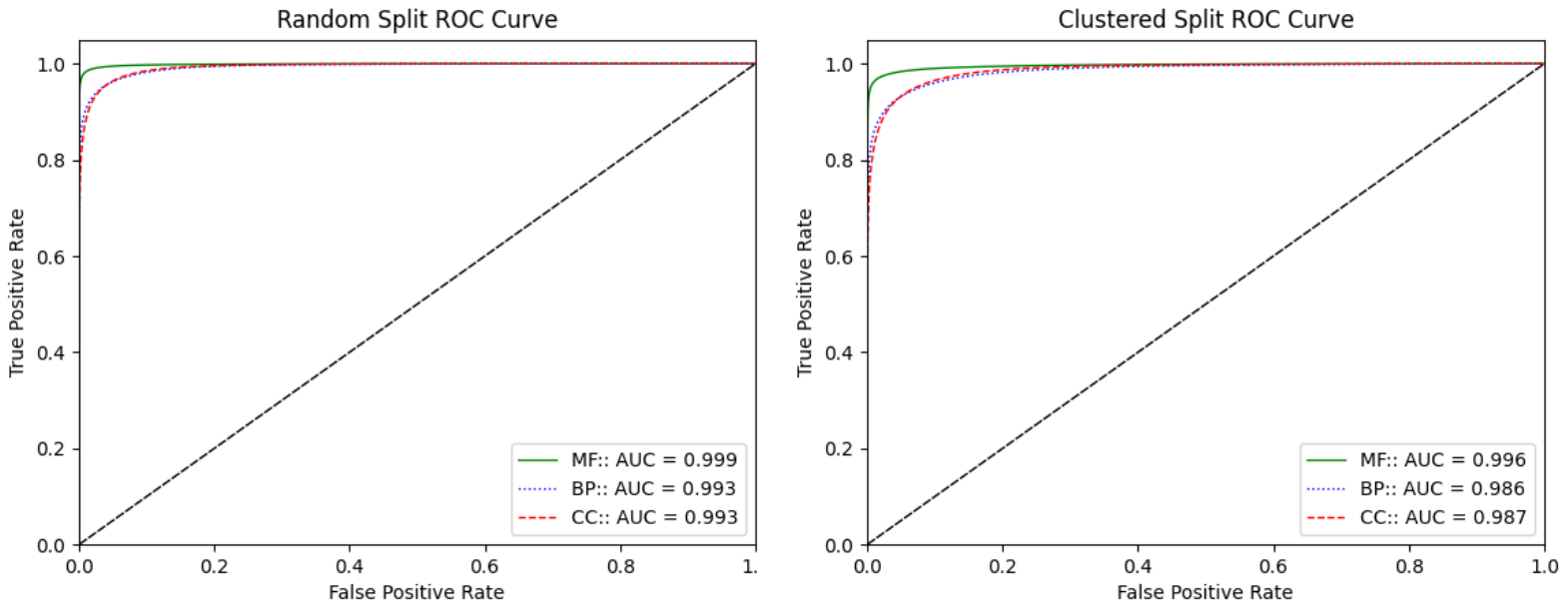}
  \caption{Left: Averaged ROC curve of the ProtGO model for the random split dataset. Right: Averaged ROC curve of the ProtGO model for the Clustered split dataset.}
  \label{Fig_ROC}
\end{figure}

\subsection{ROC Curves}
Next, the Receiver Operating Characteristic (ROC) curve is shown in Fig.\ref{Fig_ROC} for the ProtGO model on the two datasets. This plot shows the ability of a binary classifier model to differentiate between the positive and negative samples. In order to transform the ROC curve for multi-class classifications as in this study, the class wise ROC curves have been generated by considering all other classes as negative samples, followed by micro averaging all the classes to get the overall ROC curve. A straight line along the origin with a gradient of 1 represents a random classification and is shown by a black line in the figure which could to be used as a reference. while, an inverted L shaped curve touching the origin and the top two corners would represent a perfect separation between the positive and negative samples.

The ROC plots for the ProtGO model on both the random and clustered dataset splits show near perfect separation of the negative and positive samples with a large area under the curves. The model for the Molecular Function (MF) GO aspect performs the best in both the datasets with the other two closely following it. The AUC values which stands for "Area Under the Curve" is also above 99\% for the random split dataset while that for the clustered split is above 98\%. These results indicate that the proposed model is excellent at differentiating between the various classes with great ease.

\subsection{Sequence Length Analysis}

\begin{figure}[h]
  \centering
  \includegraphics[width=\linewidth]{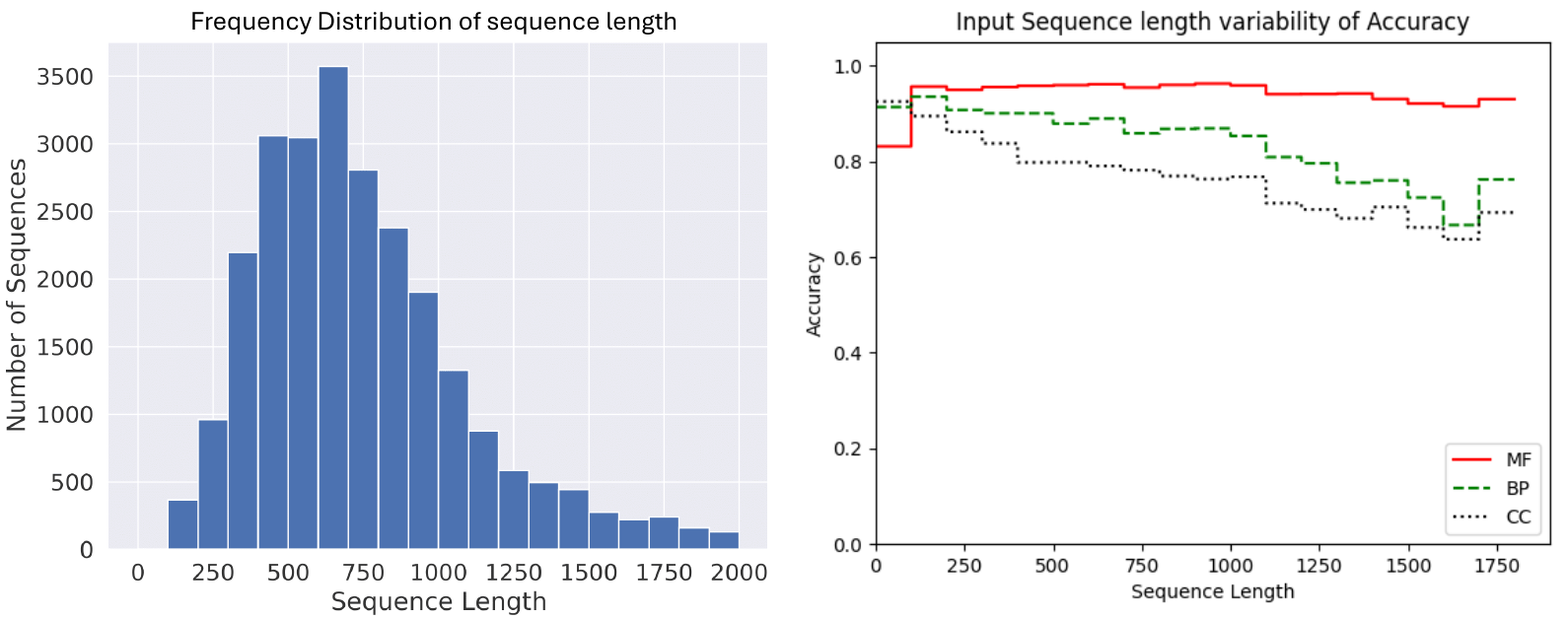}
  \caption{Left: Frequency distribution of the input protein sequence lengths in the dataset. Right: Variability of ProtGO Accuracy for the different GO aspects on the input sequence length.}
  \label{Fig_STS}
\end{figure}

In the Sequence Length Analysis (SLA) experiment, the variability of the test Accuracy on the length of the input sequence is studied along with the distribution of input sequences in the dataset. Fig. \ref{Fig_STS} shows a histogram to represent the abundance of input sequences of different lengths in the dataset along with the test Accuracy of the ProtGO model for the various input sequence lengths in the random split dataset for all three GO aspects. The frequency distribution plot peaks at around 750 tokens while most of the sequences are confined between a range of 300 to 1200 amino acids. It could be seen from the second plot in Fig. \ref{Fig_STS} that the Accuracy of the proposed model holds approximately steady till 1000 tokens while experiencing a gradual downward shift after that. This happens as the input tokens are truncated at 1000 tokens in this study for the ease of computation whereas the input sequence length could be easily increased without having to make any algorithmic changes to increase Accuracy for longer sequences. The Accuracy of the Molecular Function GO aspect does not dip even for very long sequences which shows that these GO terms are relatively easier to predict compared to the others. Whereas, the Accuracy of the other two GO aspects fall to around 70\% at very long sequence lengths.  

\section{Conclusion}
In this study, a lightweight, easy to use, transformer-based fusion machine learning model have been proposed which could predict GO terms with high Accuracy from full scale protein sequences. The performance of the model have been tested on two datasets named Random and Clustered split for all three GO aspects. The results show state-of-the-art performance on several different evaluation metrics compared to other recent related methods. The ProtGO model have been able to outperform the benchmark methods used in this study in terms of Accuracy and F1 scores by significant margins in both the random and clustered split datasets. Moreover, the results show that the margin of improvement for the ProtGO model from the benchmark methods in terms of both Accuracy and F1 scores have been significantly greater for the more challenging clustered dataset compared to the random split dataset. This indicates that the proposed architecture is better at deciphering and interpreting deep patterns and structures within the protein sequences which leads to the performance gap being greater for the more challenging dataset.

The ROC curve for the ProtGO model have also been generated for both the datasets and it shows the models ability to successfully differentiate between the positive and negative samples very effectively. This is also demonstrated by the AUC score which is above 99\% and 98\% for the random and clustered split datasets respectively. Lastly, the sequence length analysis study for the ProtGO model is also conducted where, the model demonstrating limited variability of test Accuracy as the sequence length of the input data is varied. The Accuracy is more stable upto a sequence length of a 1000 amino acids as the input sequences are truncated at that value. This indicates that the Accuracy of the model could be further increased at longer sequence lengths by raising the input sequence truncation threshold hyperparameter.

Despite several scientific studies completed recently on protein annotation utilizing deep learning models, there is a deficiency of research focused on models specifically built for GO term annotation, particularly in scenarios characterized by limited data availability. Additional efforts are necessary to gather supplementary datasets pertaining to GO term annotations to enhance the efficacy of deep learning models and improve the overall Precision of GO term predictions. Furthermore, several protein annotation models that predict distinct functionalities could be integrated into a unified framework, thereby optimizing shared computations in the analysis of protein structure. These are reserved for future endeavors.

\bibliographystyle{unsrt}  
\bibliography{main}

\begin{thebibliography}{10}

\bibitem{uniprot0}
The~UniProt Consortium.
\newblock {UniProt: a worldwide hub of protein knowledge}.
\newblock {\em Nucleic Acids Research}, 47(D1):D506--D515, 11 2018.

\bibitem{blast}
S~F Altschul, W~Gish, W~Miller, E~W Myers, and D~J Lipman.
\newblock Basic local alignment search tool.
\newblock {\em J. Mol. Biol.}, 215(3):403--410, October 1990.

\bibitem{krogh}
A~Krogh, M~Brown, I~S Mian, K~Sj{\"o}lander, and D~Haussler.
\newblock Hidden markov models in computational biology. applications to protein modeling.
\newblock {\em J. Mol. Biol.}, 235(5):1501--1531, February 1994.

\bibitem{eddy}
S~R Eddy.
\newblock {Profile hidden Markov models.}
\newblock {\em Bioinformatics}, 14(9):755--763, 10 1998.

\bibitem{Interpro}
Matthias Blum, Hsin-Yu Chang, Sara Chuguransky, Tiago Grego, Swaathi Kandasaamy, Alex Mitchell, Gift Nuka, Typhaine Paysan-Lafosse, Matloob Qureshi, Shriya Raj, Lorna Richardson, Gustavo~A Salazar, Lowri Williams, Peer Bork, Alan Bridge, Julian Gough, Daniel~H Haft, Ivica Letunic, Aron Marchler-Bauer, Huaiyu Mi, Darren~A Natale, Marco Necci, Christine~A Orengo, Arun~P Pandurangan, Catherine Rivoire, Christian J~A Sigrist, Ian Sillitoe, Narmada Thanki, Paul~D Thomas, Silvio C~E Tosatto, Cathy~H Wu, Alex Bateman, and Robert~D Finn.
\newblock The {InterPro} protein families and domains database: 20 years on.
\newblock {\em Nucleic Acids Res.}, 49(D1):D344--D354, January 2021.

\bibitem{pfam}
Sara El-Gebali, Jaina Mistry, Alex Bateman, Sean~R Eddy, Aurélien Luciani, Simon~C Potter, Matloob Qureshi, Lorna~J Richardson, Gustavo~A Salazar, Alfredo Smart, Erik~L L Sonnhammer, Layla Hirsh, Lisanna Paladin, Damiano Piovesan, Silvio~C E Tosatto, and Robert~D Finn.
\newblock {The Pfam protein families database in 2019}.
\newblock {\em Nucleic Acids Research}, 47(D1):D427--D432, 10 2018.

\bibitem{surveyTL}
Karl Weiss, Taghi~M. Khoshgoftaar, and DingDing Wang.
\newblock A survey of transfer learning.
\newblock {\em Journal of Big Data}, 3(1):9, May 2016.

\bibitem{deepGo}
Maxat Kulmanov, Mohammed~Asif Khan, and Robert Hoehndorf.
\newblock {DeepGO: predicting protein functions from sequence and interactions using a deep ontology-aware classifier}.
\newblock {\em Bioinformatics}, 34(4):660--668, 10 2017.

\bibitem{ECpred}
Alperen Dalkiran, Ahmet~Sureyya Rifaioglu, Maria~Jesus Martin, Rengul Cetin-Atalay, Volkan Atalay, and Tunca Do{\u{g}}an.
\newblock Ecpred: a tool for the prediction of the enzymatic functions of protein sequences based on the ec nomenclature.
\newblock {\em BMC Bioinformatics}, 19(1):334, Sep 2018.

\bibitem{ProLanGO}
Renzhi Cao, Colton Freitas, Leong Chan, Miao Sun, Haiqing Jiang, and Zhangxin Chen.
\newblock {ProLanGO}: Protein function prediction using neural machine translation based on a recurrent neural network.
\newblock {\em Molecules}, 22(10), October 2017.

\bibitem{DeepLoc}
José~Juan Almagro~Armenteros, Casper~Kaae Sønderby, Søren~Kaae Sønderby, Henrik Nielsen, and Ole Winther.
\newblock {DeepLoc: prediction of protein subcellular localization using deep learning}.
\newblock {\em Bioinformatics}, 33(21):3387--3395, 07 2017.

\bibitem{Schwartz}
Ariel~S Schwartz, Gregory~J Hannum, Zach~R Dwiel, Michael~E Smoot, Ana~R Grant, Jason~M Knight, Scott~A Becker, Jonathan~R Eads, Matthew~C LaFave, Harini Eavani, Yinyin Liu, Arjun~K Bansal, and Toby~H Richardson.
\newblock Deep semantic protein representation for annotation, discovery, and engineering.
\newblock {\em bioRxiv}, 2018.

\bibitem{DeePred}
Ahmet Sureyya~Rifaioglu, Tunca Do{\u{g}}an, Maria Jesus~Martin, Rengul Cetin-Atalay, and Volkan Atalay.
\newblock Deepred: Automated protein function prediction with multi-task feed-forward deep neural networks.
\newblock {\em Scientific Reports}, 9(1):7344, May 2019.

\bibitem{DeePre}
Yu~Li, Sheng Wang, Ramzan Umarov, Bingqing Xie, Ming Fan, Lihua Li, and Xin Gao.
\newblock {DEEPre: sequence-based enzyme EC number prediction by deep learning}.
\newblock {\em Bioinformatics}, 34(5):760--769, 10 2017.

\bibitem{DeepSF}
Jie Hou, Badri Adhikari, and Jianlin Cheng.
\newblock {DeepSF: deep convolutional neural network for mapping protein sequences to folds}.
\newblock {\em Bioinformatics}, 34(8):1295--1303, 12 2017.

\bibitem{HecNet1}
Safyan~Aman Memon, Kinaan~Aamir Khan, and Hammad Naveed.
\newblock {HECNet: a hierarchical approach to enzyme function classification using a Siamese Triplet Network}.
\newblock {\em Bioinformatics}, 36(17):4583--4589, 05 2020.

\bibitem{HECNet2}
Kinaan~Aamir Khan, Safyan~Aman Memon, and Hammad Naveed.
\newblock A hierarchical deep learning based approach for multi-functional enzyme classification.
\newblock {\em Protein Sci.}, 30(9):1935--1945, September 2021.

\bibitem{allign_free}
Riccardo Concu and M.~Natália D.~S. Cordeiro.
\newblock Alignment-free method to predict enzyme classes and subclasses.
\newblock {\em International Journal of Molecular Sciences}, 20(21), 2019.

\bibitem{mlDEEPre}
Zhenzhen Zou, Shuye Tian, Xin Gao, and Yu~Li.
\newblock mldeepre: Multi-functional enzyme function prediction with hierarchical multi-label deep learning.
\newblock {\em Frontiers in Genetics}, 9, 2019.

\bibitem{mohammed}
Mohammed AlQuraishi.
\newblock End-to-end differentiable learning of protein structure.
\newblock {\em Cell Systems}, 8(4):292--301.e3, 2019.

\bibitem{Senior2020}
Andrew~W. Senior, Richard Evans, John Jumper, James Kirkpatrick, Laurent Sifre, Tim Green, Chongli Qin, Augustin {\v{Z}}{\'i}dek, Alexander W.~R. Nelson, Alex Bridgland, Hugo Penedones, Stig Petersen, Karen Simonyan, Steve Crossan, Pushmeet Kohli, David~T. Jones, David Silver, Koray Kavukcuoglu, and Demis Hassabis.
\newblock Improved protein structure prediction using potentials from deep learning.
\newblock {\em Nature}, 577(7792):706--710, Jan 2020.

\bibitem{MSATrans}
Roshan~M Rao, Jason Liu, Robert Verkuil, Joshua Meier, John Canny, Pieter Abbeel, Tom Sercu, and Alexander Rives.
\newblock Msa transformer.
\newblock In Marina Meila and Tong Zhang, editors, {\em Proceedings of the 38th International Conference on Machine Learning}, volume 139 of {\em Proceedings of Machine Learning Research}, pages 8844--8856. PMLR, 18--24 Jul 2021.

\bibitem{Yilun}
Yilun Du, Joshua Meier, Jerry Ma, Rob Fergus, and Alexander Rives.
\newblock Energy-based models for atomic-resolution protein conformations.
\newblock {\em CoRR}, abs/2004.13167, 2020.

\bibitem{Yang}
Jianyi Yang, Ivan Anishchenko, Hahnbeom Park, Zhenling Peng, Sergey Ovchinnikov, and David Baker.
\newblock Improved protein structure prediction using predicted interresidue orientations.
\newblock {\em Proceedings of the National Academy of Sciences}, 117(3):1496--1503, 2020.

\bibitem{Biswas2021}
Surojit Biswas, Grigory Khimulya, Ethan~C. Alley, Kevin~M. Esvelt, and George~M. Church.
\newblock Low-n protein engineering with data-efficient deep learning.
\newblock {\em Nature Methods}, 18(4):389--396, Apr 2021.

\bibitem{ProGen}
Ali Madani, Bryan McCann, Nikhil Naik, Nitish~Shirish Keskar, Namrata Anand, Raphael~R. Eguchi, Po-Ssu Huang, and Richard Socher.
\newblock Progen: Language modeling for protein generation.
\newblock {\em bioRxiv}, 2020.

\bibitem{Anishchenko2021}
Ivan Anishchenko, Samuel~J. Pellock, Tamuka~M. Chidyausiku, Theresa~A. Ramelot, Sergey Ovchinnikov, Jingzhou Hao, Khushboo Bafna, Christoffer Norn, Alex Kang, Asim~K. Bera, Frank DiMaio, Lauren Carter, Cameron~M. Chow, Gaetano~T. Montelione, and David Baker.
\newblock De novo protein design by deep network hallucination.
\newblock {\em Nature}, 600(7889):547--552, Dec 2021.

\bibitem{Yang2019}
Kevin~K. Yang, Zachary Wu, and Frances~H. Arnold.
\newblock Machine-learning-guided directed evolution for protein engineering.
\newblock {\em Nature Methods}, 16(8):687--694, Aug 2019.

\bibitem{Mazurenko2020}
Stanislav Mazurenko, Zbynek Prokop, and Jiri Damborsky.
\newblock Machine learning in enzyme engineering.
\newblock {\em ACS Catalysis}, 10(2):1210--1223, Jan 2020.

\bibitem{DeepEC}
Jae~Yong Ryu, Hyun~Uk Kim, and Sang~Yup Lee.
\newblock Deep learning enables high-quality and high-throughput prediction of enzyme commission numbers.
\newblock {\em Proceedings of the National Academy of Sciences}, 116(28):13996--14001, 2019.

\bibitem{Bileschi2022}
Maxwell~L. Bileschi, David Belanger, Drew~H. Bryant, Theo Sanderson, Brandon Carter, D.~Sculley, Alex Bateman, Mark~A. DePristo, and Lucy~J. Colwell.
\newblock Using deep learning to annotate the protein universe.
\newblock {\em Nature Biotechnology}, 40(6):932--937, Jun 2022.

\bibitem{Dohan}
David Dohan, Andreea Gane, Maxwell~L. Bileschi, David Belanger, and Lucy Colwell.
\newblock Improving protein function annotation via unsupervised pre-training: Robustness, efficiency, and insights.
\newblock In {\em Proceedings of the 27th ACM SIGKDD Conference on Knowledge Discovery \& Data Mining}, KDD '21, page 2782–2791, New York, NY, USA, 2021. Association for Computing Machinery.

\bibitem{ProtTrans}
Ahmed Elnaggar, Michael Heinzinger, Christian Dallago, Ghalia Rehawi, Yu~Wang, Llion Jones, Tom Gibbs, Tamas Feher, Christoph Angerer, Martin Steinegger, Debsindhu Bhowmik, and Burkhard Rost.
\newblock {ProtTrans}: Toward understanding the language of life through self-supervised learning.
\newblock {\em IEEE Trans. Pattern Anal. Mach. Intell.}, 44(10):7112--7127, October 2022.

\bibitem{ProteinBert}
Nadav Brandes, Dan Ofer, Yam Peleg, Nadav Rappoport, and Michal Linial.
\newblock {ProteinBERT: a universal deep-learning model of protein sequence and function}.
\newblock {\em Bioinformatics}, 38(8):2102--2110, 02 2022.

\bibitem{Proteinfer}
Theo Sanderson, Maxwell~L. Bileschi, David Belanger, and Lucy~J. Colwell.
\newblock Proteinfer: deep networks for protein functional inference.
\newblock {\em bioRxiv}, 2021.

\bibitem{ProtEC}
Azwad Tamir, Milad Salem, and Jiann-Shiun Yuan.
\newblock Protec: A transformer based deep learning system for accurate annotation of enzyme commission numbers.
\newblock {\em IEEE/ACM Transactions on Computational Biology and Bioinformatics}, 20(6):3691--3702, 2023.

\bibitem{uniref}
Baris~E Suzek, Yuqi Wang, Hongzhan Huang, Peter~B McGarvey, Cathy~H Wu, and {UniProt Consortium}.
\newblock {UniRef} clusters: a comprehensive and scalable alternative for improving sequence similarity searches.
\newblock {\em Bioinformatics}, 31(6):926--932, March 2015.

\bibitem{wolf2019huggingface}
Thomas Wolf, Lysandre Debut, Victor Sanh, Julien Chaumond, Clement Delangue, Anthony Moi, Pierric Cistac, Tim Rault, R{\'e}mi Louf, Morgan Funtowicz, et~al.
\newblock Huggingface's transformers: State-of-the-art natural language processing.
\newblock {\em arXiv preprint arXiv:1910.03771}, 2019.

\bibitem{UniversalEngines}
Kevin Lu, Aditya Grover, Pieter Abbeel, and Igor Mordatch.
\newblock Pretrained transformers as universal computation engines.
\newblock {\em arXiv preprint arXiv:2103.05247}, 2021.

\bibitem{AMP_Deep}
Milad Salem, Arash Keshavarzi~Arshadi, and Jiann~Shiun Yuan.
\newblock Ampdeep: hemolytic activity prediction of antimicrobial peptides using transfer learning.
\newblock {\em BMC Bioinformatics}, 23(1):389, Sep 2022.

\end{thebibliography}

\end{document}